\documentclass[lettersize,journal]{IEEEtran}
\usepackage{amsmath,amsfonts}
\usepackage{algorithm} %
\usepackage{algpseudocode} %
\usepackage{array}
\usepackage[caption=false,font=normalsize,labelfont=sf,textfont=sf]{subfig}
\usepackage{textcomp}
\usepackage{stfloats}
\usepackage{url}
\usepackage{verbatim}
\usepackage{graphicx}
\usepackage{cite}

\usepackage[figuresright]{rotating}
\usepackage{xcolor}
\usepackage{multirow}
\usepackage{float}
\usepackage{booktabs}
\usepackage{threeparttable}
\usepackage{diagbox}

\begin{document}

\title{SHMamba: Structured Hyperbolic State Space Model for Audio-Visual Question Answering}

\author{Zhe Yang,
        Wenrui Li,
        Guanghui Cheng

\thanks{Corresponding authors: Guanghui Cheng and Wenrui Li}
\thanks{Zhe Yang and Guanghui Cheng  with the School of Mathematical Sciences, University of Electronic Science and Technology of China, Chengdu, Sichuan 611731, China. (e-mail: xiazhe200006@gmail.com; ghcheng@uestc.edu.cn).}
\thanks{Wenrui Li is with the Department of Computer Science and Technology, Harbin Institute of Technology, Harbin 150001, China. (e-mail: liwr@stu.hit.edu.cn)}
}

\markboth{Journal of \LaTeX\ Class Files,~Vol.~14, No.~10, July~2024}%
{Shell \MakeLowercase{\textit{et al.}}: A Sample Article Using IEEEtran.cls for IEEE Journals}


\maketitle

\begin{abstract}
The Audio-Visual Question Answering (AVQA) task holds significant potential for applications. Compared to traditional unimodal approaches, the multi-modal input of AVQA makes feature extraction and fusion processes more challenging. Euclidean space is difficult to effectively represent multi-dimensional relationships of data. Especially when extracting and processing data with a tree structure or hierarchical structure, Euclidean space is not suitable as an embedding space. Additionally, the self-attention mechanism in Transformers is effective in capturing the dynamic relationships between elements in a sequence. However, the self-attention mechanism's limitations in window modeling and quadratic computational complexity reduce its effectiveness in modeling long sequences. To address these limitations, we propose SHMamba: Structured Hyperbolic State Space Model to integrate the advantages of hyperbolic geometry and state space models. Specifically, SHMamba leverages the intrinsic properties of hyperbolic space to represent hierarchical structures and complex relationships in audio-visual data. Meanwhile, the state space model captures dynamic changes over time by globally modeling the entire sequence. Furthermore, we introduce an adaptive curvature hyperbolic alignment module and a cross fusion block to enhance the understanding of hierarchical structures and the dynamic exchange of cross-modal information, respectively. Extensive experiments demonstrate that SHMamba outperforms previous methods with fewer parameters and computational costs. Our learnable parameters are reduced by 78.12\%, while the average performance improves by 2.53\%. Experiments show that our method demonstrates superiority among all current major methods and is more suitable for practical application scenarios.
\end{abstract}

\begin{IEEEkeywords}
Audio-Visual learning, Question Answering, Multi-modal Learning
\end{IEEEkeywords}

\section{Introduction}
\IEEEPARstart{W}ith the development and establishment of the Audio-Visual Question Answering (AVQA) datasets\cite{b11},\cite{b12}, the AVQA field has become an important branch of multi-modal research in recent years\cite{b7},\cite{b8},\cite{b9}. AVQA combines the features of Visual Question Answering (VQA)\cite{b1},\cite{b2},\cite{b3},\cite{b4} and Audio Question Answering (AQA)\cite{b5},\cite{b6}, focusing on understanding and analyzing audio-visual signals to answer relevant questions. The AVQA task focus on utilizing audio and visual information, to understand complex audio-visual scenes more comprehensively and generate accurate answers to specific questions. This capability shows great potential in various applications, such as automated video content analysis, human-computer interaction, and so on. Especially in human-computer interaction, AVQA helps machines understand multi-modal commands and automatically recognize and respond to complex speech, facial expressions, and gestures, making interactions more natural and efficient. By further analyzing audio-visual content, AVQA systems provide detailed interpretations of scenes, characters, actions, and emotions, promoting intelligent media management and content creation with higher levels of automation and personalization.

\begin{figure}
  \centering
  \subfloat[ ]{\includegraphics[width=0.45\linewidth]{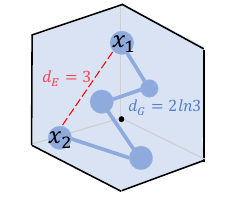}\label{fig:sub1}}
  \hfill 
  \subfloat[ ]{\includegraphics[width=0.45\linewidth]{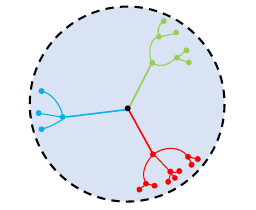}\label{fig:sub2}}
  \caption{In (a), the Euclidean distance and the graph distance between \(x_1\) and \(x_2\) are \(3\) and \(2\ln_{}{3} \), respectively. (b) is a visualization of a 2-D Poincaré ball.}
  \label{fig:images}
\end{figure}
The goal of AVQA is to understand and answer complex questions about content that integrates audio and visual elements. Although significant research achievements have been made\cite{b8},\cite{b9}, many challenges still exist. \textbf{1) Representation Learning of Audio-Visual Data}: Audio-visual data contain complex hierarchical structures that are closely integrated into continuous video and audio streams. This indicates that effective information extraction requires going beyond surface-level audio and visual features. When processing these multi-level pieces of information, previous AVQA models\cite{b7},\cite{b8},\cite{b9} focus on data representation and processing in non-curved Euclidean spaces. However, Euclidean space faces challenges when dealing with datasets that have complex structures, such as hierarchical or tree-like data\cite{b10},\cite{b13},\cite{b14},\cite{b48}. Embedding such data in Euclidean space can lead to significant distortions\cite{b10}. As shown in Fig. \ref{fig:images}(a), where the distance \(d_E\) between nodes $x_1$ and $x_2$ in Euclidean space is \(3\), but their actual distance \(d_G\) in the graph's structure is \(2\ln_{}{3} \), the Euclidean distance fails to accurately reflect their true relationship. In this case, hyperbolic geometry demonstrates its advantages. As demonstrated in the Poincaré ball model\cite{b46} in Fig. \ref{fig:images}(b), hyperbolic space more naturally accommodates such hierarchical data. In hyperbolic space, the way distances are calculated better preserves the inherent structure of the data, reducing distortion and more accurately representing the actual relationships between data points. \textbf{2) Computational Complexity}: The AVQA task involves audio and visual comprehension in long videos, indicating a large amount of information and computational cost in processing this data. Current research primarily utilizes transformer-based models, which capture complex interactions and dependencies within sequences through their self-attention mechanisms. Although many works\cite{b25},\cite{b26},\cite{b27} have demonstrated the superiority of transformer models in capturing fine-grained features, they still face potential weaknesses when dealing with long sequence data. The self-attention mechanism in transformers struggles to effectively focus on content beyond a fixed window size. Additionally, its computational complexity increases quadratically with the sequence length. These factors result in inefficiency when dealing with long video sequences.

\begin{figure*}
    \centering
    \includegraphics[width=0.9\linewidth]{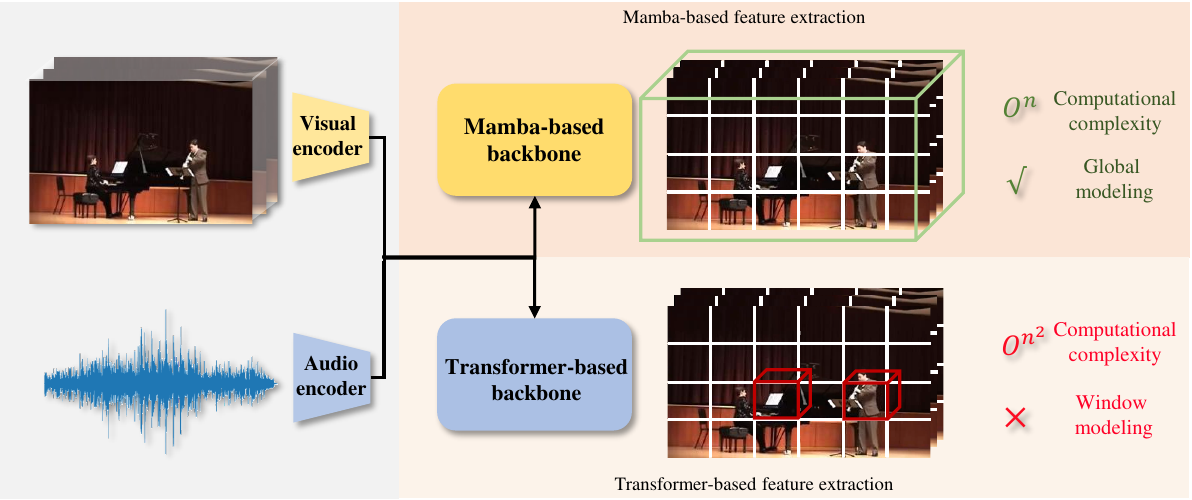}
    \caption{When processing long sequence information, networks based on Transformer differ from those based on Mamba. The Mamba-based method can model the entire sequence and is more efficient in handling long sequence data.} 
    \label{fig1}
\end{figure*}
To tackle these challenges, we propose SHMamba: Structured Hyperbolic State Space Model. SHMamba leverages the intrinsic properties of hyperbolic space to represent the hierarchical structures and complex relationships between audio-visual data, while the State Space Model (SSM) captures the dynamic changes in the temporal dimensions of the video. By combining the advantages of hyperbolic and SSM, SHMamba significantly improves the processing of complex audio-visual structures in long video sequences.

Specifically, our model uses three linear layers to extract semantic information from audio, visual, and question inputs. Due to its exponentially growing volume, hyperbolic space helps distribute embeddings in a tree-like structure. We project the extracted audio-visual embeddings into hyperbolic space to capture the internal hierarchical structure of the data and learn more discriminative embeddings. To reduce the computational cost of long sequence data, we adopt the structured state space model proposed by mamba\cite{b41}, which effectively addresses the computational inefficiency of transformer models on long sequence data while remaining global video information (See Fig. \ref{fig1}). Finally, we utilize the cross fusion block to enhance the interaction of cross-modal features.

Our main contributions are summarized as follows:
\begin{itemize}
\item We propose SHMamba to enhance performance in the AVQA task. By integrating hyperbolic geometry with a structured SSM, our approach focuses on the complicated structure of audio-visual content while modelling global information. To our knowledge, this is the first work to combine hyperbolic geometry and SSM.

\item We introduce the hyperbolic alignment module (HAM) with adaptive curvature and the cross fusion block (CFB). HAM utilizes adaptive curvature parameters to flexibly explore the inherent hierarchical structure of data for different data characteristics and organization. Simultaneously, CFB integrates multi-modal information by utilizing the dynamic information interaction.

\item Experimental results show that our model outperforms previous methods while reducing the number of learnable parameters. Our ablation study demonstrates the importance of the proposed model and its components.
\end{itemize}

In Section \ref{A}, we discuss the related work. We introduce the background in audio-visual question answering, hyperbolic geometry, and state-space models. Section \ref{B} introduces our proposed SHMamba model and related algorithms. Section \ref{C} shows the model's effectiveness with experimental results and visualizations. Finally, in Section \ref{D} we summarise the research results, emphasising the main findings and contributions.


\section{Related Work}
\label{A}
\subsection{Audio-Visual Question Answering} 
The AVQA tasks are intended to improve comprehension of complex scenes by combining audio and visual inputs. Recently, many studies develop excellent model frameworks for the AVQA task. Li et al.\cite{b28} proposes a method that combines spatial and temporal localization modules. This method establishes cross-modal associations between audio and visual data and focuses on key time segments. Nadeem et al.\cite{b29} reduces spatial misalignment with parameter-free context blocks. They use self-supervised learning for audio-visual temporal alignment and achieve effective balancing and semantic processing of audio-visual information through a chain of cross-attention modules. Chen et al.\cite{b30} employ a dual-stage framework to obtain both global and local features from audio-visual information. These features are then integrated under the direction of question-aware guidance. These methods provide new perspectives for further exploring features in audio-visual data.

In this work, we propose a new multi-modal learning framework. This architecture captures the hierarchical structure within audio-visual content more effectively and enhances the understanding of long-sequence data.

\subsection{Hyperbolic Geometry} 
Hyperbolic geometry becomes an effective method for constructing structured representations due to its natural advantage in encoding hierarchical structures\cite{b10}. It is widely applied across various areas, such as metric learning, and graph data analysis\cite{b35},\cite{b36}. In the field of vision, hyperbolic embedding enhances tasks such as semantic segmentation\cite{b31}, anomaly recognition\cite{b32}, and action recognition\cite{b33}, demonstrating its potential for capturing the hierarchical structure of visual data. The AVQA task requires integrating audio and visual information to answer questions about complex multi-modal scenes. Audio-visual data possesses a rich hierarchical structure. For instance, the AVQA dataset, which is based on the VGGSound dataset\cite{b34} as the source of its original video data, categorizes all ``sound" types in VGGSound into eight parent categories: ``Animals," ``Nature," ``Humans," ``Machines," ``Musical Instruments," ``Home and Daily Life," ``Environment," and ``Others." This hierarchical classification reflects the complex organizational structure of audio-visual data. Existing AVQA methods focus on finding cross-modal correspondences between audio-visual streams in Euclidean space, overlooking the potential hierarchical structures. In contrast, hyperbolic space offers a unique perspective to handle the intrinsic hierarchy, which is crucial for finely understanding and answering questions based on complex scenes.

In this paper, we model the embeddings of audio and visual modalities in hyperbolic space. Our model elevates the hierarchy and multi-modal associations in the AVQA task to a new level by mapping and aligning audio and visual data in hyperbolic space. Additionally, we design an adaptive curvature module on the hyperbolic manifold to perform feature alignment, avoiding the situation where a fixed curvature is unsuitable for the complex structure of audio-visual data.

\subsection{State Space Model} 
State Space Models (SSM) demonstrate their advantages in modeling long sequence data in tasks such as computer vision, natural language processing and video understanding\cite{b38},\cite{b37},\cite{b39},\cite{b40},\cite{b50}. Structured State Space Sequence models (S4) and their variants, such as Mamba\cite{b41}, address the limitations of traditional recurrent neural networks in long sequence modeling by optimizing complexity and improving sequence processing speed\cite{b42},\cite{b43}. Notably, SSMs implement autoregressive reasoning in a recursive manner while remaining parallel processing capabilities. This combination provides unique advantages in both efficiency and performance. The AVQA task involves long video sequences containing complex temporal dependencies and cross-modal content. Previous AVQA work\cite{b7},\cite{b8},\cite{b9} uses Transformers as the underlying structure and achieves good results in local attention. However, the transformer is unable to model the content outside the window and suffers from secondary computation problems. Performance bottlenecks and computational resource limitations are encountered when processing long video sequences and performing cross-modal fusion.

We consider using State Space Models (SSMs) for temporal modeling to provide more precise control over long-term dependencies and cross-modal interactions. Our model overcomes the challenge of capturing global information in long video content, offering a more effective way to simulate dynamic changes in video sequences and interactions between audio and video elements. In addition, to better integrate audio and video information, we introduce cross fusion blocks to enhance the interaction between the two modalities. This multimodal fusion not only helps in more accurate information retrieval, but further provides richer contextual information for complex question and answer tasks.
\begin{figure*}
	\centering
	\includegraphics[width=1\linewidth]{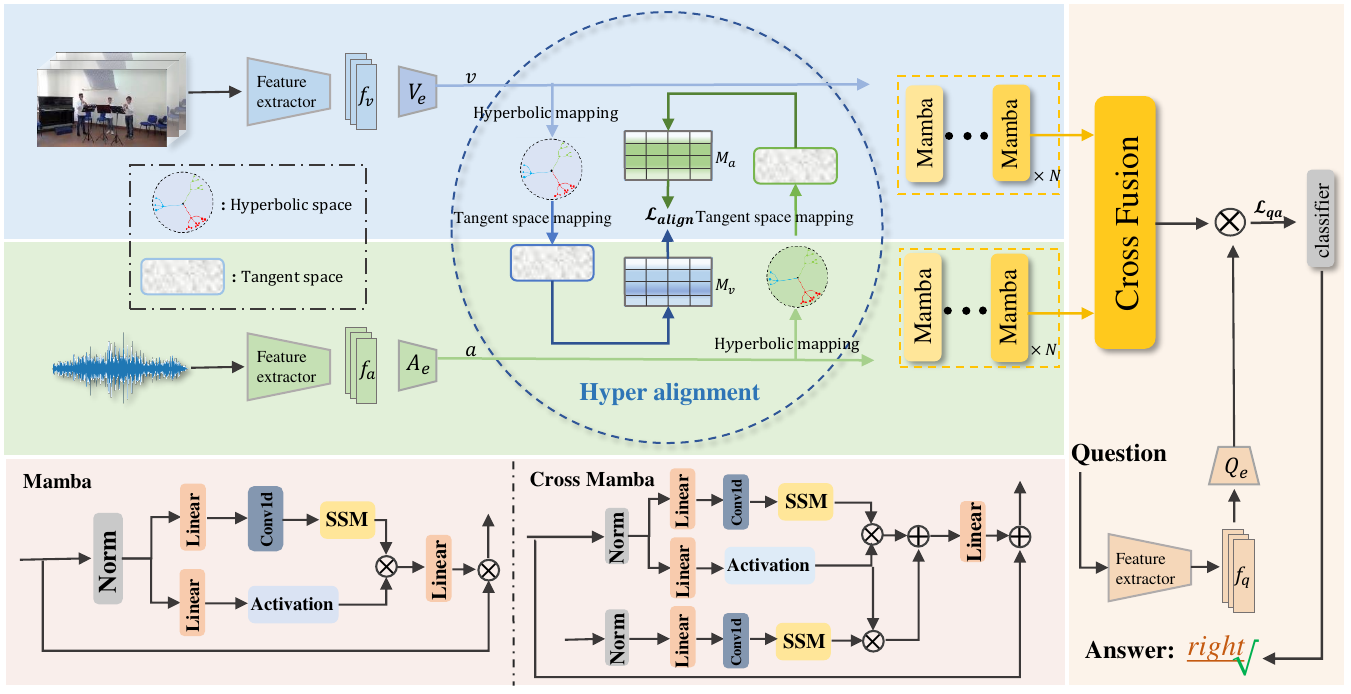}
	\caption{Architecture of the SHMamba. First, we employ a pre-trained model to extract features from audio, visual, and question. Next, we simply use the three linear layers as encoders for audio, vision, and problem, respectively. We align audio and visual features by hyperbolic mapping to a hyperbolic space. At the same time, we capture the dynamics of spatio-temporal changes within the video via the Mamba module. Finally the audio and visual features are further interacted and fused through the cross fusion block to predict the answers to the input questions.}
	\label{fig2}
\end{figure*}

\section{Methodology}
\label{B}
To address the Audio-Visual Question Answering (AVQA) problem, we propose SHMamba: Structured Hyperbolic State Space Model, which focuses on the hierarchical structure of audio-visual content and understands changes in content over time. Compared to single-mode networks, SHMamba can better understand scene content. Fig. \ref{fig2} provides an overview of the proposed framework.
\subsection{Input Representation} 

For each input video sequence containing audio and visual data, we split it into \(L\) pairs of corresponding audio and visual segments \(\{A_l, V_l\}_{l=1}^L\), each lasting \(T\) seconds. We extract audio features \(f_a\) from each audio segment \(a_t\) using the pre-trained VGGish model\cite{b44}, where \(f_a \in R^{T \times D}\). For visual feature extraction, we uniformly sample the same number of frames in all video clips. Then, we use the pre-trained CLIP model\cite{b45} to extract the visual features \(f_v\) for each video segment \(V_t\), where \(f_v \in R^{T \times D}\). For the posed question \(Q = \{q_k\}_{k=1}^K\), we convert each word \(q_k\) into a vector representation of fixed length. These vectors are then fed into the pre-trained CLIP model\cite{b45} to get the overall feature \(f_q\) of the question, where \(f_q \in R^{D}\).

\subsection{Hyperbolic Space Alignment} 
\subsubsection{Background} Hyperbolic space is a curved space with constant negative curvature ( \(k < 0\)). To achieve hyperbolic alignment, we need to project embedding points from Euclidean space into hyperbolic space. This process includes two stages: hyperbolic mapping and tangent space mapping.

\textbf{Hyperbolic Mapping.} In this paper, we use the Poincaré ball model\cite{b46} to represent hyperbolic space. We define an \(n\)-dimensional Poincaré ball with curvature \(k\) as follows:

\begin{equation}
\begin{aligned}
\mathbb{P}_k^n = \{ x \in \mathbb{R}^n : \|x\| < 1/|k| \},\
\end{aligned}
\end{equation}
where \(\|\cdot\|\) denotes the Frobenius norm.

To project a point \(x \in \mathbb{R}^n\) from Euclidean space \(\mathbb{R}^n\) into an \(n\)-dimensional Poincaré ball \(\mathbb{P}_k^n\) with curvature \(k\), we perform the following hyperbolic mapping:

\begin{equation}
\begin{aligned}
\
x_P = \Gamma_\mathbb{P}(x) = 
\begin{cases} 
x, & \text{if } \|x\| \leq 1/|k| \\
\frac{1 - \varepsilon }{|k|} \frac{x}{\|x\|}, & \text{else}
\end{cases}
\
\end{aligned}
\label{e2}
\end{equation}
where \(x_P\) is the projected point in the Poincaré sphere and \(\varepsilon\) is a scaling factor adjusted according to the curvature \(k\).

\textbf{Tangent Space Mapping.} The hyperbolic tangent space is a local Euclidean space that approximates the hyperbolic space. To facilitate the calculation of loss, we project points from hyperbolic space into the hyperbolic tangent space. Given a point \( z_P \in \mathbb{P}_k^n \) in the Poincaré sphere, we can generate a tangent plane \(T_z\mathbb{P}_k^n\):

\begin{equation}
\begin{split}
z_P \oplus_k x_P &= \frac{(1 + 2|k| \langle z_P, x_P \rangle + |k| \|x_P\|^2) z_P}
{1 + 2|k| \langle z_P, x_P \rangle + |k|^2 \|z_P\|^2 \|x_P\|^2} \\
&\quad + \frac{(1 - |k| \|z_P\|^2) x_P}
{1 + 2|k| \langle z_P, x_P \rangle + |k|^2 \|z_P\|^2 \|x_P\|^2},
\end{split}
\end{equation}

\begin{equation}
\begin{aligned}
x_{Tg} =  \frac{2}{\sqrt{|k| \lambda_k(z_P)}} &\tanh^{-1}\left(\sqrt{|k|} \|  - z_P \oplus_k x_P \| \right) \\
& \cdot \frac{-z_P \oplus_k x_P}{\| -z_P \oplus_k x_P \|},
\end{aligned}
\label{e4}
\end{equation}
where \(\langle , \rangle\) denotes the inner product and \(x_{Tg}\in T_z\mathbb{P}_k^n\). To facilitate understanding and analysis, we set \( z_P = 0_P \) .
\subsubsection{Hyperbolic alignment loss}
Our goal is to use alignment loss to enhance the hierarchical structure reflected in the audio-visual features \(v\) and \(a\). Since it is not clear which modality contributes more to hierarchy exploration, our model ensures alignment between features by minimizing differences in similarity within each modality.

To construct the loss function, we first project the features \(v\) and \(a\) of the two modalities to points \(P_v\) and \(P_a\) in the Poincaré ball space using Eq. \eqref{e2}. We then project \(P_v\) and \(P_a\) from hyperbolic space to the tangent space using Eq. \eqref{e4}, obtaining \(T_v\) and \(T_a\). Essentially, we compute the mean element-wise squared difference between two similarity matrices. The features of the two modalities are aligned by reducing the distance between their similarities. The hyperbolic alignment loss \(\mathcal{L}_{align}\) is formulated as follows:

\begin{equation}
\begin{aligned}
\
\mathcal{L}_{align} = \|W_{v, norm} - W_{a, norm}\|^2
,\
\end{aligned}
\end{equation}
where \(W_{v, norm}\) and \(W_{a, norm} \in \mathbb{R}^{B \times B}\) are the normalized matrices of the intramodal similarity matrices \(W_v\) and \(W_a \in \mathbb{R}^{B \times B}\) for a batch of visual or audio features \(T_v\) and \(T_a\) in the hyperbolic tangent space. Each element of matrix \(W\) is obtained by computing the cosine similarity across various features in the tangent space.

The use of fixed curvature may not effectively address the complex structure of audio-visual data, which may affect the quality of the representation of audiovisual features. To solve this problem, we introduce the concept of using adaptive curvature for feature alignment on hyperbolic manifolds. Specifically, the modal features \(v\) and \(a\) are first concatenated into a vector \(K_{av}\), and then the curvature is computed through a linear layer (MLP). The specific methods are as follows:

\begin{equation}
\begin{aligned}
\ k = k_0 \cdot \text{sigmoid}(\text{MLP}(K_{av})) .\
\end{aligned}
\label{e6}
\end{equation}
where \( \text{MLP}(\cdot) \) denotes a fully connected layer, and \( k_0 \) represents the initial curvature.

\subsection{Structured Spatial Model}
\subsubsection{Background}

The structured spatial model represents the dynamics of a system as the evolution of a hidden state, which is connected to the input and output through a series of linear relationships. Specifically, a linear SSM can be represented by the following combination of ordinary differential equations:
\begin{equation}
\begin{aligned}
\begin{cases}
h'(t) = Ah(t) + Bx(t), \\y(t) = Ch(t),
\end{cases}
\end{aligned}
\label{e7}
\end{equation}
where $h(t) \in \mathbb{R}^N$ denotes the hidden state vector, $x(t) \in \mathbb{R}^L$ represents the input vector, and $y(t) \in \mathbb{R}^L$ is the output vector. The system's dynamics are defined by three key matrices: $A \in \mathbb{R}^{N \times N}$, the state transition matrix, which models the evolution of the hidden states; $B \in \mathbb{R}^{N \times L}$, the input matrix, which projects the input vector into the state space; and $C \in \mathbb{R}^{L \times N}$, the output matrix, which projects the state space back to the output space. These matrices collectively capture the linear relationships between the states, inputs, and outputs, forming the backbone of the SSM's predictive framework.

To adapt to the digital computational environment, continuous SSMs need to be discretized. The S4 and mamba models\cite{b41} achieve this by applying the Zero-Order Hold method, which converts the continuous system into its discrete form:

\begin{equation}
\begin{aligned}
\begin{cases}
 \bar{A} = \exp(\Delta A),  \\
\bar{B} = (\Delta A)^{-1}(\exp(\Delta A) - I) \cdot \Delta B, \\
h_t = \bar{A}h_{t-1} + \bar{B}x_t, \\
y_t = C h_t.
\end{cases}
\end{aligned}
\end{equation}
SSMs provide a clear description of system dynamics and enable efficient computation using existing digital signal processing tools. See Algorithm \ref{alg1} for Mamba block details.

\subsubsection{Cross Fusion Block}
Inspired by the idea of cross-attention\cite{b47}, our work introduce a cross fusion block to enhance the interaction and integration of cross-modal features. We project the features from both modalities into a shared space. Then, We use a gating mechanism to encourage learning of complementary features and suppress redundant features. In addition, to enhance the local features, we add a deep convolution function, which improves the coding ability of local features in the fusion process. The details of the cross fusion block process are detailed in Algorithm \ref{alg2}.

\begin{algorithm}[t]
\caption{Mamba Block}
\label{alg1}
\begin{algorithmic}[1] 
\fontsize{9}{10}\selectfont 
\Require token sequence \(T_{l-1}\) : \((B,N,C)\) .
            \Ensure token sequence \(T_l\) : \((B,N,C)\).
                \State \hfill \textcolor{gray}{/***  \(Normalize\) \(the\) \(input\) \(sequence\) ***/}
                \State \(T'_{l-1}:(B,N,C)=Layernorm(T_{l-1})\)
                \State \(x:(B,N,M)=Linear^x(T'_{l-1})\)
                \State \(z:(B,N,M)=Linear^z(T'_{l-1})\)
                \State \hfill \textcolor{gray}{/***  \(Parameters\) \(function\) ***/}
                \State \(x':(B,N,M)=SiLU(Conv1d(x))\)
                \State \(B:(B,N,C) \longleftarrow Linear^B(x')\)
                \State \(C:(B,N,C) \longleftarrow Linear^C(x')\)
                \State \(\Delta :(B,N,M) \longleftarrow log(1 + exp(Linear^\Delta  (x') + Parameter^\Delta  ))\)
                \State \(\bar{A} : (B, N, M, L) \longleftarrow  \Delta  \otimes  Parameter^A\)
                \State \(\bar{B} : (B, N, M, L) \longleftarrow  \Delta  \otimes  B\)
                \State \(y:(B,N,M)=SSM(\bar{A},\bar{B},C)(x')\)
                \State \hfill \textcolor{gray}{/***  \(Get\) \(gated\) \(y'\) ***/}
            \State \(y':(B,N,M)=y\odot SiLU(z)\)
            \State \(T_l:(B,N,C)=Linear(y')+T_{l-1}\)
\end{algorithmic}
\end{algorithm}

\begin{figure*}[t]
    \resizebox{\textwidth}{!}{
        \begin{minipage}{\textwidth}
            \begin{algorithm}[H]
            \caption{Coss Fusion Block}
            \label{alg2}
            \begin{algorithmic}[1] 
            \fontsize{10}{12}\selectfont 
            \Require token sequence \(T^a_{l-1}\):\((B,N,C)\) , Visual feature \(T^v_{l-1}\):\((B,N,C)\).
            \Ensure token sequence \(T^a_{l}\):\((B,N,C)\) ,  \(T^v_{l}\):\((B,N,C)\).
            \For{each modality $m$ in \{$a$, $v$\}} 
\State  \hfill \textcolor{gray}{/*** \(Normalize\) \(the\) \(input\) \(sequence\) ***/}
                \State \(T'^{m}_{l-1}:(B,N,C)=Layernorm(T^{m}_{l-1})\)
                \State \(x_m:(B,N,M)=Linear^m(T'^{m}_{l-1})\)
                \State  \hfill \textcolor{gray}{/*** \(Parameters\) \(function\) ***/}
                \State \(x'_m:(B,N,M)=SiLU(Conv1d(x_m))\)
                \State \(B_m:(B,N,C) \longleftarrow Linear^B(x'_m)\)
                \State \(C_m:(B,N,C) \longleftarrow Linear^C(x'_m)\)
                \State \(\Delta_m :(B,N,M) \longleftarrow  log(1 + exp(Linear^\Delta  (x'_m) + Parameter^\Delta  ))\)
                \State \(\bar{A} : (B, N, M, L) \longleftarrow  \Delta _m \otimes  Parameter^A\)
                \State \(\bar{B} : (B, N, M, L) \longleftarrow  \Delta _m \otimes  B\)
                \State \(y_{m}=SSM(\bar{A}_m,\bar{B}_m,C_m)(x'_m)\)
            \EndFor
                \State  \hfill \textcolor{gray}{/***  \(Get\) \(gated\) \(y'_{a}\) \(and\) \(y'_{v}\) ***/}
                \State \(z=Linear^v(T'^{v}_{l-1})\)
                \State \(y'_{a}=y_{a}\odot SiLU(z)\)
                \State \(y'_{v}=y_{v}\odot SiLU(z)\)

            \State \(T^{a}_l=Linear^a(y'_{a}+y'_{v})+T^{a}_{l-1}\)
            \State \(T^{v}_l=Linear^v(y'_{a}+y'_{v})+T^{v}_{l-1}\)
            \end{algorithmic}
            \end{algorithm}
        \end{minipage}
    }
\end{figure*}

\subsection{Answer Prediciton and Loss Function}
To address audio-visual question answering,  we predict the answer to a specific question using the combined multi-modal embedding \( f_{av} \). Following the methodology outlined\cite{b24}, we select a correct word from a predefined answer vocabulary as the answer. Specifically, our approach consists of applying a linear layer and a softmax function to compute the probability of a candidate answer \( p \in \mathbb{R}^C \). The predicted probability vector and the corresponding true label \( y \) allow us to compute the cross-entropy loss:

\begin{equation}
\begin{aligned}
\ \mathcal{L}_{qa} = -\sum_{c=1}^C y_c \log(p_c) .\
\end{aligned}
\end{equation}

The total loss \( \mathcal{L} \) can be expressed as follows:
\begin{equation}
\begin{aligned}
\ \mathcal{L} = \mathcal{L}_{align} + \mathcal{L}_{qa}.\
\end{aligned}
\end{equation}
This combination of loss functions, including hyperbolic alignment loss and cross-entropy loss, enables the model to better understand and distinguish complex hierarchical structures inherent in data. By aligning audio and visual features in hyperbolic space, the model minimizes distortion typically seen in Euclidean spaces and enhances the representation of complex data relationships. Additionally, cross-entropy loss helps in accurately predicting answers to AVQA tasks. The total loss function takes advantage of hyperbolic geometry and SSM to significantly improve the performance of the multi-modal question answering task.

\begin{table*}
    \centering
    \begin{threeparttable}
        \caption{The performance comparison on MUSIC-AVQA datasets.}
        \label{TAB1}
        \fontsize{9}{14}\selectfont 
        \setlength{\tabcolsep}{3pt} 
        \begin{tabular}{c|c|ccc|ccc|cccccc|c}
            \toprule[1.5pt]  
            \multirow{2}{*}{\textbf{Task}} &
            \multirow{2}{*}{\textbf{Method}} &
            \multicolumn{3}{c|}{\begin{tabular}[c]{@{}c@{}}\textbf{Audio}\end{tabular}} &
            \multicolumn{3}{c|}{\begin{tabular}[c]{@{}c@{}}\textbf{Visual}\end{tabular}} &
            \multicolumn{6}{c|}{\begin{tabular}[c]{@{}c@{}}\textbf{Audio-Visual}\end{tabular}} &
            \multirow{2}{*}{\textbf{Avg(\%)}} {\begin{tabular}[c]{@{}c@{}}\end{tabular}} \\ 
            && Count & Comp & Avg  & Count & Local & Avg  & Exist & Count & Local & Comp & Temp & Avg \\ \hline
            
        \multirow{2}{*}{\textbf{AudioQA}} & FCNLSTM\cite{b5}\textsuperscript{TASLP2019}     & 70.80 & 65.66 & 68.90 & 64.58 & 48.08 & 56.23 & 82.29 & 59.92 & 46.20 & 62.94 & 47.45 & 60.42 & 60.81 \\ 
        & CONVLSTM \cite{b5}\textsuperscript{TASLP2019}       & 74.07 & \textbf{68.89} & 72.15 & 67.47 & 54.56 & 60.94 & 82.91 & 50.81 & 63.03 & 60.27 & 51.58 & 62.24 & 63.65\\
          \hline
        \multirow{3}{*}{\textbf{VisualQA}} & GRU\cite{b16}\textsuperscript{ICCV2015}         & 71.29 & 63.13 & 68.28 & 66.08 & 68.08 & 67.09 & 80.67 & 61.03 & 51.74 & 62.85 & 57.79 & 63.03 & 65.03 \\
        & HCAttn\cite{b17}\textsuperscript{NeurIPS2016}       & 70.80 & 54.71 & 64.87 & 63.49 & 67.10 & 65.32 & 79.48 & 59.84 & 48.80 & 56.31 & 56.33 & 60.32 & 62.45\\
        & MCAN\cite{b18}\textsuperscript{CVPR2019}     & 78.07 & 57.74 & 70.58 & 71.76 & 71.76 & 71.76 & 80.77 & 65.22 & 54.57 & 56.77 & 46.84 & 61.52 & 65.83\\
                    \hline
        \multirow{3}{*}{\textbf{VideoQA}} & PSAC\cite{b19}\textsuperscript{AAAI2019}        & 75.02 & 66.84 & 72.00 & 68.00 & 70.78 & 69.41 & 79.76 & 61.66 & 55.22 & 61.13 & 59.85 & 63.60 & 66.62\\
        & HME\cite{b20}\textsuperscript{CVPR2019}   & 73.65 & 63.74 & 69.89 & 67.42 & 70.20 & 68.83 & 80.87 & 63.64 & 54.89 & 63.03 & 60.58 & 64.78 & 66.75\\
        & HCRN\cite{b21}\textsuperscript{CVPR2020}        & 71.29 & 50.67 & 63.69 & 65.33 & 64.98 & 65.15 & 54.15 & 53.28 & 41.74 & 51.04 & 46.72 & 49.82 & 56.34\\
          \hline
        \multirow{4}{*}{\textbf{AVQA}}& AVSD\cite{b22}\textsuperscript{CVPR2019}        & 72.41 & 61.90 & 68.52 & 67.39 & 74.19 & 70.83 & 81.61 & 58.79 & 63.89 & 61.52 & 61.41 & 65.49 & 67.44\\
        & Pano-AVQA\cite{b23}\textsuperscript{ICCV2021}       & 75.71 & 65.99 & 72.13 & 70.51 & 75.76 & 73.16 & 82.09 & 65.38 & 61.30 & 63.67 & 62.04 & 66.97 & 69.53\\ 
        & ST-AVQA\cite{b24}\textsuperscript{CVPR2022 }       & 78.18 & 67.05 & 74.06 & 71.56 & 76.38 & 74.00 & 81.81 & 64.51 & \textbf{70.80} & \textbf{66.01} & 63.23 & 69.54 & 71.52\\
        \cline{2-15}
            & \textbf{SHMamba (ours)} & \textbf{82.30} & 63.64 & \textbf{75.42} & \textbf{78.53} & \textbf{81.31} &\textbf{ 79.93} & \textbf{82.89} & \textbf{72.65} & 67.93 & 61.31 & \textbf{68.37} & \textbf{70.64} & \textbf{74.12}\\
            \bottomrule[1.5pt]
        \end{tabular}
    \end{threeparttable}
\end{table*}

\begin{table*}
    \centering
    \begin{threeparttable}
        \caption{Ablation results on the MUSCI-AVQA dataset.}
        \label{TAB2}
        \fontsize{9}{13}\selectfont 
        \setlength{\tabcolsep}{4.5pt} 
        \begin{tabular}{c|ccc|ccc|cccccc|c}
            \toprule[1.5pt]  
            \multirow{2}{*}{\textbf{Model }} &
            \multicolumn{3}{c|}{\begin{tabular}[c]{@{}c@{}}\textbf{Audio}\end{tabular}} &
            \multicolumn{3}{c|}{\begin{tabular}[c]{@{}c@{}}\textbf{Visual}\end{tabular}} &
            \multicolumn{6}{c|}{\begin{tabular}[c]{@{}c@{}}\textbf{Audio-Visual}\end{tabular}} &
            \multirow{2}{*}{\textbf{Avg(\%)}} {\begin{tabular}[c]{@{}c@{}}\end{tabular}} \\ 
            & Count & Comp & Avg  & Count & Local & Avg  & Exist & Count & Local & Comp & Temp & Avg \\ \hline

         W/o Mamba Block & 76.79 & 58.75 & 70.14 & 74.02 & 79.43 & 76.75 & 82.09 & 69.72 & 60.98 & 61.31 & 65.21 & 67.99 & 70.70 \\ W/o Cross Fuison Block   & 77.68 &58.25 & 70.52 & 75.69 & 80.16 & 77.95 & 83.20 & 71.23 & 64.57 & 58.31 & 66.67 & 68.78 & 71.52 \\ 
         W/o Hyper Alignment   
         & 80.73 & 61.28 & 73.56 & 76.78 & 79.92 & 78.36 & 82.29 & 73.60 & 63.80 & 62.40 & 66.30 & 69.92 & 72.90 \\
                   \hline
         \textbf{SHMamba} & 82.30 & 63.64 & \textbf{75.42} & 78.53 & 81.31 & \textbf{79.93} & 82.89 & 72.65 & 67.93 & 61.31 & 68.37 & \textbf{70.64} & \textbf{74.12}\\

            \bottomrule[1.5pt]
        \end{tabular}
    \end{threeparttable}
\end{table*}

\section{Experiments}
\label{C}

This section offers a comprehensive performance evaluation of our SHMamba model. We begin by detailing our experimental settings and datasets. Subsequently, we compare our results with prior methods, emphasizing SHMamba's superior capabilities. We also provide an analysis of learnable parameters and floating-point operations (FLOPs), demonstrating the efficiency of our model compared to leading methods. An ablation study further analyses the impact of specific components and settings, validating their importance in the model's performance. In addition, we provide t-SNE visualization results to represent the effect of the distribution of hyperbolic space in the hierarchy of audio-visual data. These visualizations enhance our quantitative analysis by demonstrating SHMamba's practical effectiveness in real-life scenarios. Lastly, we discuss the model’s limitations, particularly in integrating the advantages of hyperbolic space with sequence modeling, setting the stage for future enhancements.

\subsection{Datasets and Experimental Settings}
We conduct our study on two AVQA datasets: MUSIC-AVQA\cite{b11} and AVQA\cite{b12}. The MUSIC-AVQA dataset consists of 9,288 videos primarily in a concert setting and 45,867 Q\&A pairs. These videos cover 22 different musical instruments, totaling over 150 hours of content. The AVQA dataset contains 57,015 videos reflecting everyday life scenarios and 57,335 Q\&A pairs, which rely on audio and visual clues and cover eight types of questions, including ``Which," ``Where," and ``Why." For both datasets, we use the official benchmark division, splitting them into training, evaluation, and test sets.

We use the method described in \cite{b24} for feature extraction. The SHMamba model trains on 6 Nvidia V100 GPUs. For the SHMamba model, the dimension of the hidden states in the linear transformation is set to 256, and the dropout rate is set to 0.1. Additionally, we set the hyperparameter \(k_0\) shown in Eq. \eqref{e6} to -0.1. And the number of audio and visual Mamba blocks \(N\) are both set to 4. Our SHMamba model is trained using the Adam optimizer. The initial learning rate was set to 0.0001. We set the batch size to 32 and the maximum number of training epochs to 30.

\begin{table}[t]
	\centering
	\begin{threeparttable}
		\caption{Learnable parameter and FLOPs of ST-AVQA and SHMamba.}
		\label{TAB3}
		\fontsize{9}{14}\selectfont 
		\setlength{\tabcolsep}{6pt} 
		\begin{tabular}{c|cc|c}  
			\toprule[1.5pt]  
			Method &  
			Learnable Param (M) &
			FLOPs (G) & Avg(\%)\\ 
			\hline
			ST-AVQA & 18.480 & 3.188 & 71.52 \\
 \hline
			SHMamba & \textbf{4.033} & \textbf{1.086}& \textbf{74.12}  \\ 
			\bottomrule[1.5pt]
		\end{tabular}
	\end{threeparttable}
\end{table}

\begin{table}[t]
	\centering
	\begin{threeparttable}
		\caption{The performance comparison on AVQA datasets.}
		\label{TAB4}
		\fontsize{9}{14}\selectfont 
		\setlength{\tabcolsep}{12pt} 
		\begin{tabular}{c|c|c}  
			\toprule[1.5pt]  
			Method &  
			Ensemble  & Total Accuracy(\%)\\ 
			\hline
            LADNet & HAVF   & 84.1 \\
			HME & HAVF   & 85.0 \\
			PSAC & HAVF & 87.4  \\ 
   		HGA & HAVF & 87.7  \\ 
			ACRTransformer & HAVF & 87.8  \\ 
			HCRN & HAVF & 89.0  \\ 
			\textbf{SHMamba} & -- & \textbf{90.8}  \\ 
			\bottomrule[1.5pt]
		\end{tabular}
	\end{threeparttable}
\end{table}

\begin{figure}
	\centering
	\includegraphics[width=1.0\linewidth]{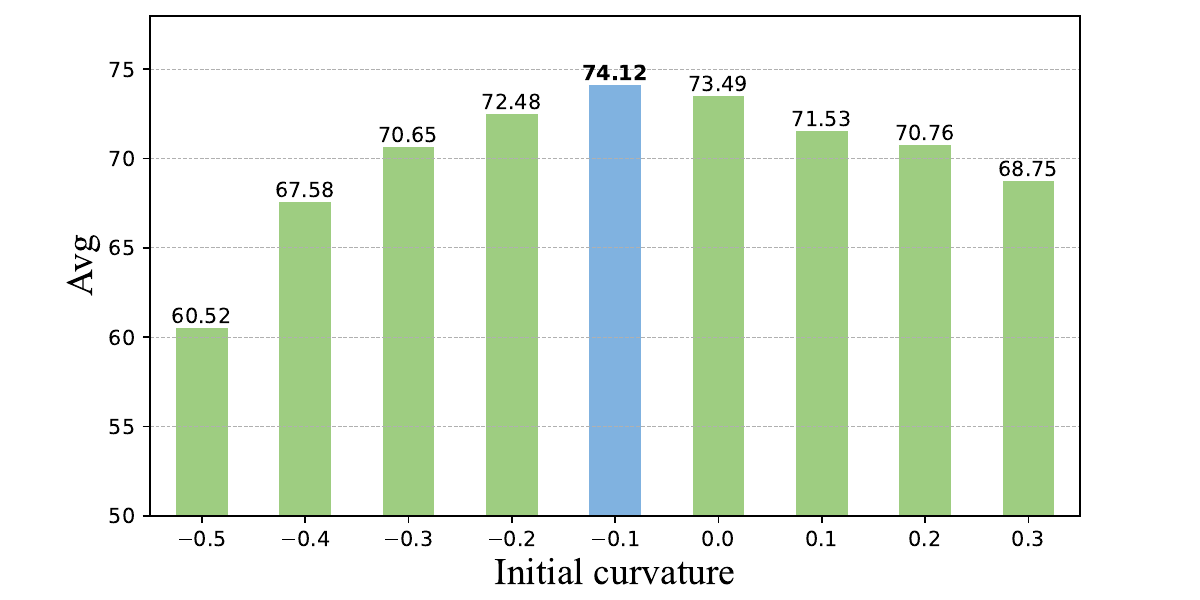}
	\caption{Ablation Study of curvature value.}
	\label{fig5}
\end{figure}

\begin{figure}
	\centering
	\includegraphics[width=1\linewidth]{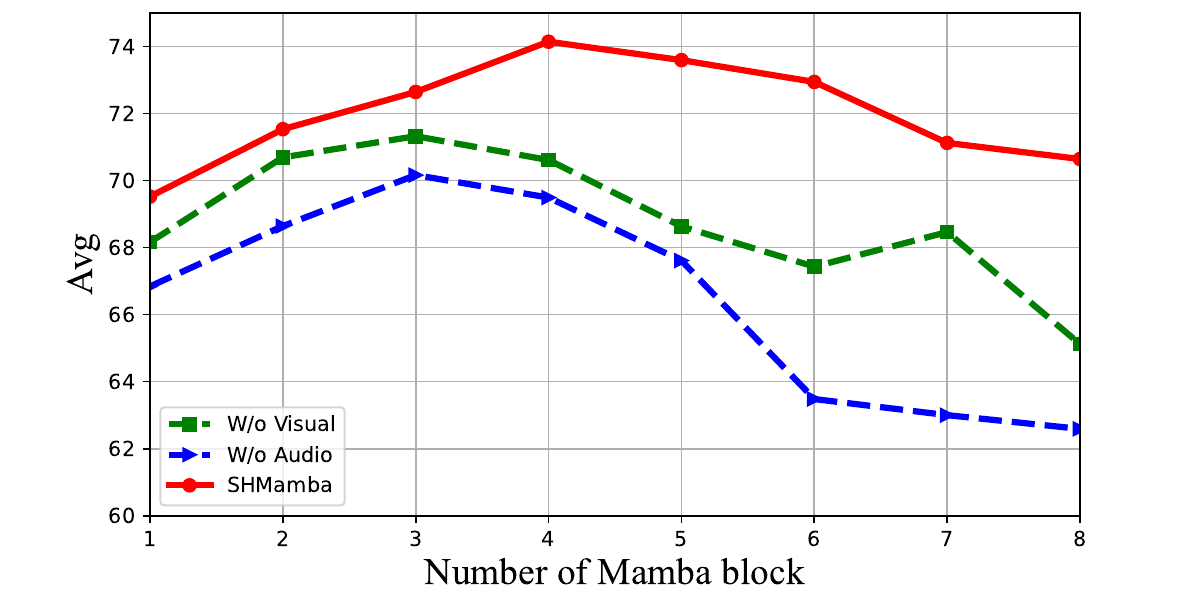}
	\caption{Ablation study of Mamba modules.}
	\label{fig6}
\end{figure}
\subsection{Results and Comparison}
\subsubsection{Results on the MUSIC-AVQA dataset}
We compare our SHMamba model with recent baselines on MUSIC-AVQA, including FCNLSTM\cite{b5}, GRU\cite{b16}, HCAttn\cite{b17}, MCAN\cite{b18}, PSAC\cite{b19}, HME\cite{b20}, HCRN\cite{b21}, CONVLSTM\cite{b5}, AVSD\cite{b22}, Pano-AVQA\cite{b23}, ST-AVQA\cite{b24}. Table \ref{TAB1} shows that our SHMamba achieves superior performance, outperforming most baselines. Compared to the previously best-performing ST-AVQA, SHMamba demonstrates exceptional performance in all aspects. Specifically, our method overall improves by 2.60\%, with average increases of 1.36\%, 5.93\%, and 1.1\% in Audio, Visual, and Audio-Visual scenarios, respectively. These results indicate that SHMamba has a greater ability to recognize and handle audio-visual scenes in a variety of ways. This may be due to the SHMamba model's ability to leverage the properties of hyperbolic space to explore more hierarchical in audio-visual data, combined with the model's excellent temporal processing capabilities.

\subsubsection{Results on the AVQA dataset}
We evaluate the robustness of our proposed model under various conditions by conducting experiments on the AVQA dataset. This dataset consists of diverse audio-visual question-answering tasks. It involves complex scenes and question types and comprehensively testing the model's generalization ability and stability. The results are summarized in Table \ref{TAB3}. The table shows that our model outperforms previous models in all metrics, clearly indicating the significant potential of SHMamba. Specifically, the SHMamba model enhances the understanding of subtle semantic differences in video content and speech questions by introducing structured hyperbolic mapping. This is particularly important when dealing with question-answer pairs that contain ambiguous contexts and complex semantic relationships.
\begin{figure*}
	\centering	\includegraphics[width=0.9\linewidth]{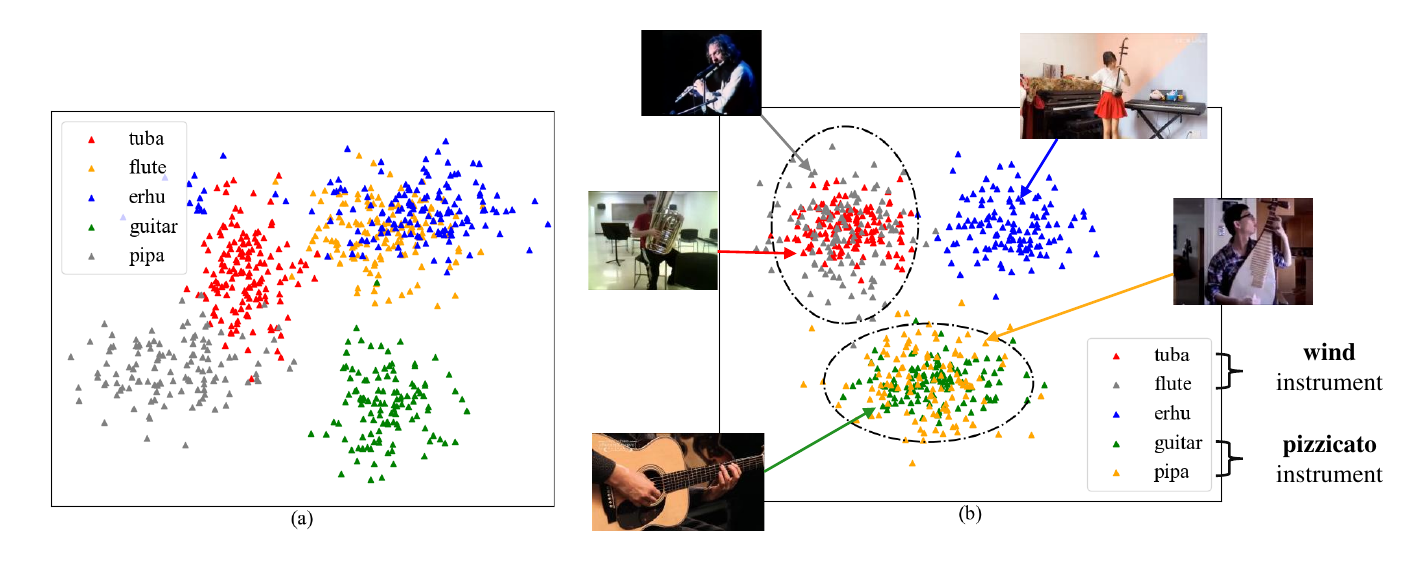}
	\caption{Visualization examples on MUSIC-AVQA. We give t-SNE visualization results for five categories of hyperbolic aligned features: `tuba', `flute', `erhu', `guitar' and `pipa'. They can be categorized into two parent classes: ``wind instrument" and ``pizzicato instrument". SHMamba can learn hyperbolic alignment loss to align features of the same parent class.}
	\label{fig3}
\end{figure*}
\subsection{Ablation Study}
In this section, we conduct an ablation study of the key components of the model to improve our understanding of SHMamba. The results are shown in Table \ref{TAB2}. Our model consists of two main parts: 1) exploring the deep semantic relationship between vision and audio through the hyperbolic alignment module, and 2) using a structured SSM to explore the temporal and spatial features of long sequences.

First, we remove the hyperbolic alignment module to explore the alignment effect of deep semantic features between audio and vision. As shown in Table \ref{TAB2} ``W/o Hyperbolic Alignment", the performance decreases by 1.22\% when the model only has shallow features. Next, we perform the same operation on the Mamba module. After losing the Mamba module's capability to capture temporal and spatial features, the accuracy drops by 3.42\%. The results show that removing the Mamba module leads to a decline in model performance, proving its effectiveness. Finally, we perform ablation on the cross fusion block. We replace the cross fusion block with a concat operation for the fusion of the two modal features. Clearly, as shown in Table \ref{TAB2}, Our cross fusion block is crucial for facilitating feature interaction and integration across different modalities. We speculate that this module achieves the effect of reducing redundant features and effectively integrating audio and visual features.

\subsection{Quantitative Results}
In Table \ref{TAB2}, we use two metrics, learnable parameters and floating point operations (FLOPs), to demonstrate the efficiency of our proposed SHMamba model and compare them with ST-AVQA. The comparison results show that SHMamba significantly reduces learnable parameters (18.48M compared to 4.03M) and FLOPs (3.188G compared to 1.086G), with reductions of 78.12\% and 65.93\%, respectively. This situation is due to the reason that they both use a transformer as the backbone of the model. The internal self-attention mechanism of transformers captures complex interactions and dependencies within sequences. However, each layer of a transformer has independent parameters. Particularly for videos, the long-sequence characteristics significantly increase the computation volume due to the quadratic calculation of self-attention, leading to a significant increase in model parameters and FLOPs. In contrast, SHMamba shares the same linear dynamic parameters across multiple time steps through its unique structured state-space model. This suggests that it only needs to focus on some key information in the sequence. Consequently, it reduces the number of parameters and FLOPs, effectively minimizing redundancy. Importantly, this decrease in computational demands also improves accuracy, highlighting our model's efficiency. As Table \ref{TAB2} illustrates, SHMamba achieves excellent performance with fewer parameters and lower FLOPs, demonstrating its outstanding efficiency. In summary, compared to existing methods, SHMamba offers significant improvements in computational efficiency.

\subsection{Hyper Parameter Setting}
\subsubsection{Curvature}
We conduct experiments at different curvatures to understand the impact of hyperbolic geometric distortion on model performance in Euclidean space. The model's performance is tested across various curvatures, with the results shown in Fig. \ref{fig5}. The optimal performance is achieved at a curvature of -0.1, indicating that this value provides the best hyperbolic alignment effect.
\subsubsection{Number of Mamba blocks}
We conduct experiments on the number of Mamba modules \( N \) under the conditions of ``W/o Visual", ``W/o Audio", and the complete model. (Note: ``W/o Visual" denotes the number of Visual Mamba blocks is 0, and ``W/o Audio" is similar.) The results are shown in Fig. \ref{fig6}. When extracting temporal and spatial features from a single modality, the performance is best when \( N = 3 \). This may be due to the lack of sufficient temporal and spatial information when only one modality is used. The performance is optimal for the complete model when \( N = 4 \), indicating that the Mamba blocks' ability to extract temporal and spatial information is maximized at this point.

\begin{figure*}
	\centering
	\includegraphics[width=1.0\linewidth]{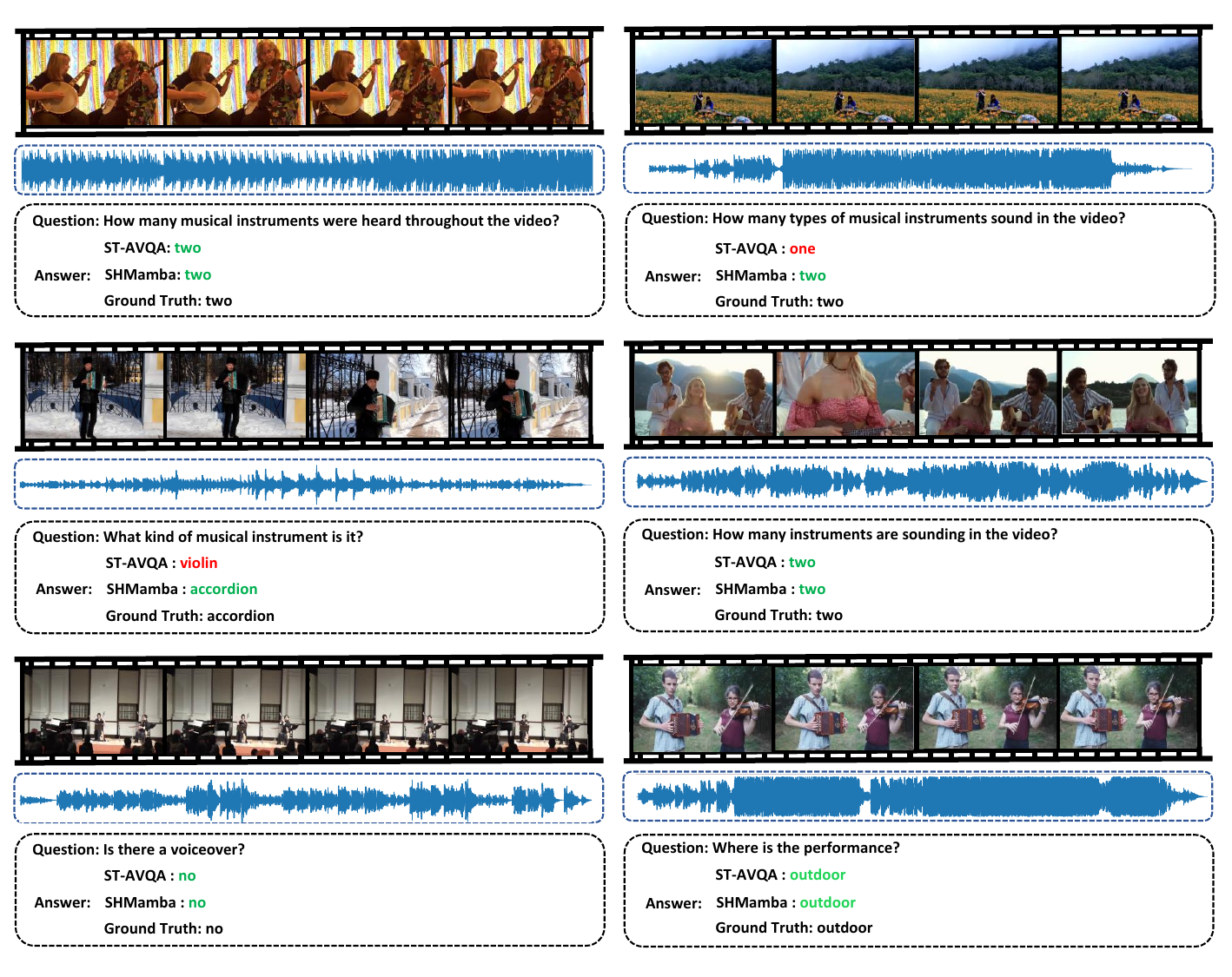}
	\caption{Visual comparison of AVQA on MUSIC-AVQA. Correct answers are marked in green and incorrect answers are marked in red.}
	\label{fig4}
\end{figure*}
\subsection{Visualization}
\subsubsection{t-SNE Visualizations}
We use t-SNE visualization results to demonstrate the capability of hyperbolic space in exploring data hierarchies. As shown in Fig. \ref{fig3}. In the MUSIC-AVQA dataset, hyperbolic alignment loss actively pulls features from the same parent category closer together and pulls features from different parent categories farther apart. For example, the features of ``tuba" and ``flute" belonging to the same parent category of ``wind instruments" come closer together, while the features of ``guitar" and ``lute" belonging to the category of ``plucked instruments" separate. This demonstrates that aligning audio and visual features in hyperbolic space through a hyperbolic method enhances the accuracy of audio-visual question answering.

\subsubsection{Visualization of Question Answer}
In Fig. \ref{fig4}, we present our method's results and compare them with the ST-AVQA method and the ground truth. The results demonstrate that our method outperforms the ST-AVQA. The ST-AVQA method predicts the incorrect answer ``one", while our method accurately predicts ``two". We use both audio and visual information to determine the spatial location of the ``instrument". Our method effectively captures and integrates spatial and temporal dimensions, and semantically associates sounds and appearances with the ``instrument category".

\subsection{Limitation}
Our model effectively explores audio-visual features in terms of geometry and spatio-temporal aspects. However, it remains challenging to combine the advantages of hyperbolic space in spatial information exploration with those of SSMs in long sequence modeling.

\section{Conclusion}
\label{D}
In this paper, we proposed a novel multi-modal framework for audio-visual question answering tasks. Our framework integrated geometric space and structured SSMs, enhancing the interaction and fusion of audio and visual features across temporal and spatial dimensions. We conducted experiments on the hyperparameters of the proposed modules to optimize the model's performance. Additionally, to demonstrate the effectiveness and superiority of our model, we performed ablation analysis on various components of the model and compared the visualized results with other models.

\bibliographystyle{unsrt}
\bibliography{refs}

\vfill

\begin{IEEEbiography}[{\includegraphics[width=1in,height=1.25in,clip,keepaspectratio]{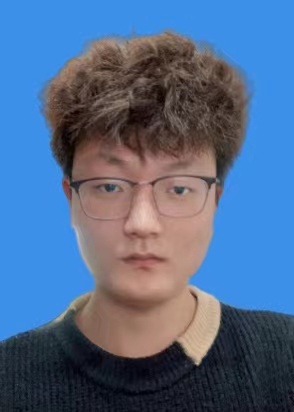}}]{Zhe Yang}  received the B.S. degree from the College of Science, Northeast Forestry University, China, in 2022. He is currently pursuing the M.E. degree with the University of Electronic Science and Technology of China. His research interests include zero-shot learning, audio-visual learning, and computer vision.
\end{IEEEbiography}

\begin{IEEEbiography}[{\includegraphics[width=1in,height=1.25in,clip,keepaspectratio]{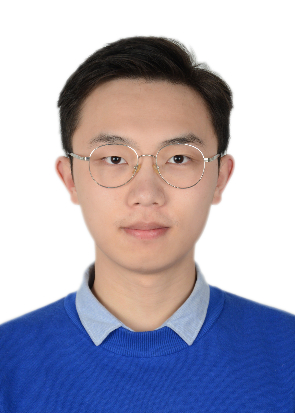}}]{Wenrui Li} received the B.S. degree from the School of Information and Software Engineering, University of Electronic Science and Technology of China (UESTC), Chengdu, China, in 2021. He is currently working toward the Ph.D. degree from the School of Computer Science, Harbin Institute of Technology (HIT), Harbin, China. His research interests include multimedia search, joint source-channel coding, and spiking neural network. He has authored or co-authored more than 15 technical articles in referred international journals and conferences. He also serves as a reviewer for IEEE TCSVT, IEEE TMM, NeurIPS, ECCV, AAAI, and ACM MM.
\end{IEEEbiography}

\begin{IEEEbiography}[{\includegraphics[width=1in,height=1.25in,clip,keepaspectratio]{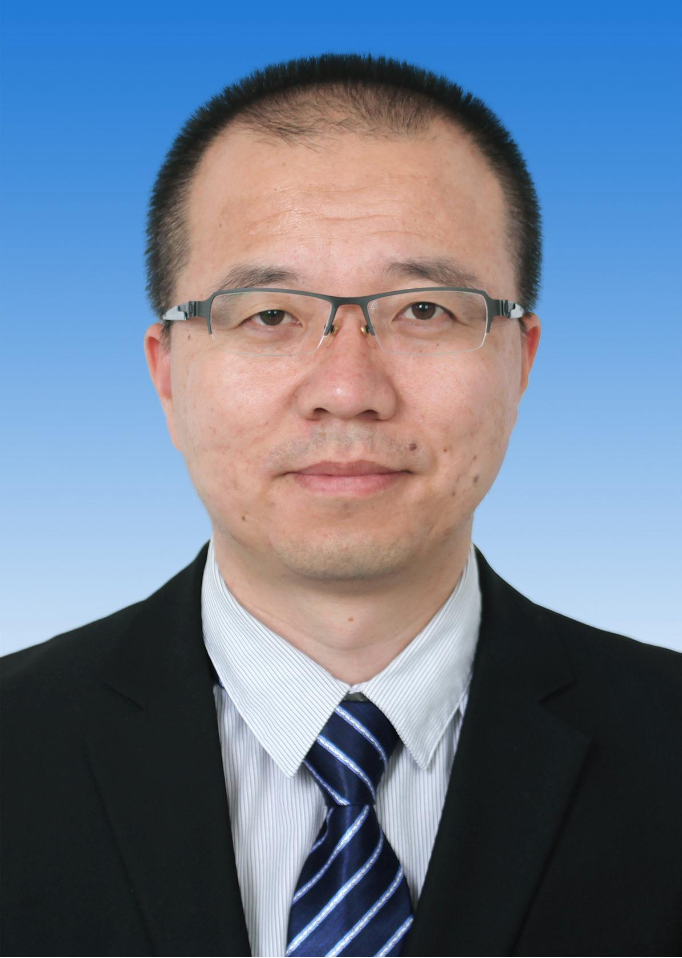}}]{Guanghui Cheng}  was born in Jilin, China, in 1979. He received the Ph.D. degree in applied mathematics from University of Electronic Science and Technology of China, Sichuan, China, in 2008. He is currently a professor with University of Electronic Science and Technology of China, Sichuan, China. His current research interests include matrix computation, matrix and tensor decomposition, signal processing and image processing. He has authored or co-authored more than 50 technical articles in referred international journals and conferences.
\end{IEEEbiography}
\end{document}